\documentclass[conference]{IEEEtran}
\IEEEoverridecommandlockouts
\usepackage{cite}
\usepackage{amsmath,amssymb,amsfonts}
\usepackage{algorithmic}
\usepackage[dvipdfmx]{graphicx}
\usepackage{textcomp}
\usepackage{xcolor}
\usepackage{comment}
\usepackage{multirow}
\usepackage[caption=false]{subfig}
\usepackage{tabularx}
\usepackage[super]{nth}

\def\BibTeX{{\rm B\kern-.05em{\sc i\kern-.025em b}\kern-.08em
    T\kern-.1667em\lower.7ex\hbox{E}\kern-.125emX}}
    
\begin{document}
\bstctlcite{IEEEexample:BSTcontrol}
\title{
Leveraging IoT and Weather Conditions to Estimate the Riders Waiting for the Bus Transit on Campus \\ 
\thanks{\textsuperscript{1} Ismail Arai was also affiliated as a visiting scholar at James Madison University, Virginia, USA while working on parts of this paper.}
}

\author{
\IEEEauthorblockN{Ismail Arai \textsuperscript{1}}
\IEEEauthorblockA{Information Initiative Center\\Nara Institute of Science and Technology\\ 
Nara, Japan\\
ismail@itc.naist.jp}
 \and
\IEEEauthorblockN{Ahmed Elnoshokaty}
\IEEEauthorblockA{
Computer Information Systems\\ 
Northern Michigan University\\
Michigan, USA\\
aelnosho@nmu.edu}
\and
\IEEEauthorblockN{Samy El-Tawab}
\IEEEauthorblockA{College of Integrated Science and Engineering\\
James Madison University\\
Virginia, USA\\
 eltawass@jmu.edu}
 }

\maketitle
\begin{abstract}

The communication technology revolution in this era has increased the use of smartphones in the world of transportation. In this paper, we propose to leverage IoT device data, capturing passengers' smartphones' Wi-Fi data in conjunction with weather conditions to predict the expected number of passengers waiting at a bus stop at a specific time using deep learning models. Our study collected data from the transit bus system at James Madison University (JMU) in Virginia, USA. This paper studies the correlation between the number of passengers waiting at bus stops and weather conditions. Empirically, an experiment with several bus stops in JMU, was utilized to confirm a high precision level. We compared our Deep Neural Network (DNN) model against two baseline models: Linear Regression (LR) and a Wide Neural Network (WNN). The gap between the baseline models and DNN was 35\% and 14\% better Mean Squared Error (MSE) scores for predictions in favor of the DNN compared to LR and WNN, respectively.

\end{abstract}


\section{Introduction}\label{sec:intro}
For the Transit Bus Management System, data-driven fleet management strategies empowered by precise models that predict the number of passengers waiting at the bus stops are essential. On a university campus, students often have to wait in additional time when a packed bus arrives. To solve the problem, we have developed the Internet of Things (IoT) system~\cite{el2017data} that estimates the number of passengers by analyzing passengers' smartphones' Wi-Fi frames. Even though the system can grasp the bus stop's occasion, it still has not been spun out from a research prototype with the lack of feasibility in the real field. Other data sources captured from the bus (e.g., camera feed) can be considered ground truth \cite{ryu2020wifi}. 

This paper proposes a machine learning utilization technique that employs the IoT passenger counter system to provide the labeled training data. The learning phase inputs the weather data, transportation information, campus schedule, and the number of passengers waiting at the bus stop. After building the machine learning model, it can estimate the number of passengers without the IoT system. Our group has noticed a significant relationship between the number of passengers waiting for the bus at a particular bus station and the weather conditions. Using these machine learning models, we can achieve a realistic fleet management system without having IoT sensors installed permanently at each bus stop.

The experiment took place in public transit systems at James Madison University (JMU) in Harrisonburg, Virginia, USA, in Spring 2017. With a month of data collected at seven bus stops, the Mean Squared Error (MSE) of Deep Neural Network (DNN), Wide Neural Network (WNN), and Linear Regression (LR) were 1.15, 1.34, 1.77, respectively. As a result of the experiment, we achieved the best MSE with the DNN.



\section{Related Work}\label{sec:related}

Several researchers have used IoT in the intelligent transportation world. Applications inside cities such as Smart Parking, Bus Monitoring have been easier to deploy with the integration of IoT devices to sense data~\cite{khanna2016iot, garcia2016secure}. At the same time, other highway applications (e.g., highway monitoring and incident detection) advanced with the use of IoT devices~\cite{salem2015driveblue,popescu2017automatic}. The IoT devices' role has improved with the advance of power batteries and reliability of IoT~\cite{chand2018survey}. Our system depends on IoT devices to improve the public transit system's quality of service~\cite{el2020framework}. Several researchers have integrated the power of communication and public transportation to improve service quality (e.g., ridership and waiting time in public bus transit) ~\cite{dunlap2016estimation, oransirikul2016feasibility}. Dunlap et al.~\cite{dunlap2016estimation} estimated passenger origin and destination (OD) information for transit lines using IoT sensors to collect Wi-Fi and Bluetooth beacon.

The rise of data-mining-based studies and machine learning techniques has recently improved research quality in many fields. In Intelligent Transportation, Lathia et al. proposed a machine learning technique to enhance the the passenger's ticket choice by studying travel history patterns and mining the public transport fare data collected from the bus system~\cite{lathia2013individuals}. On the other hand, Amato et al. studied car parking occupancy with deep learning techniques~\cite{amato2016car}. Tahere et al. predicted five crowding levels with rich data such as ridership 15 minutes ago~\cite{Tahereh2019}. Machine learning techniques and deep learning will open the door toward more improvement to big data and data-driven systems.

In this paper, we leverage machine learning models to study bus ridership in one of the USA's college cities. We integrate several parameters such as class schedule and weather information to predict hourly passengers waiting at bus stops.

\begin{figure}[!ht]
\centering
 \includegraphics[width=\linewidth,clip]{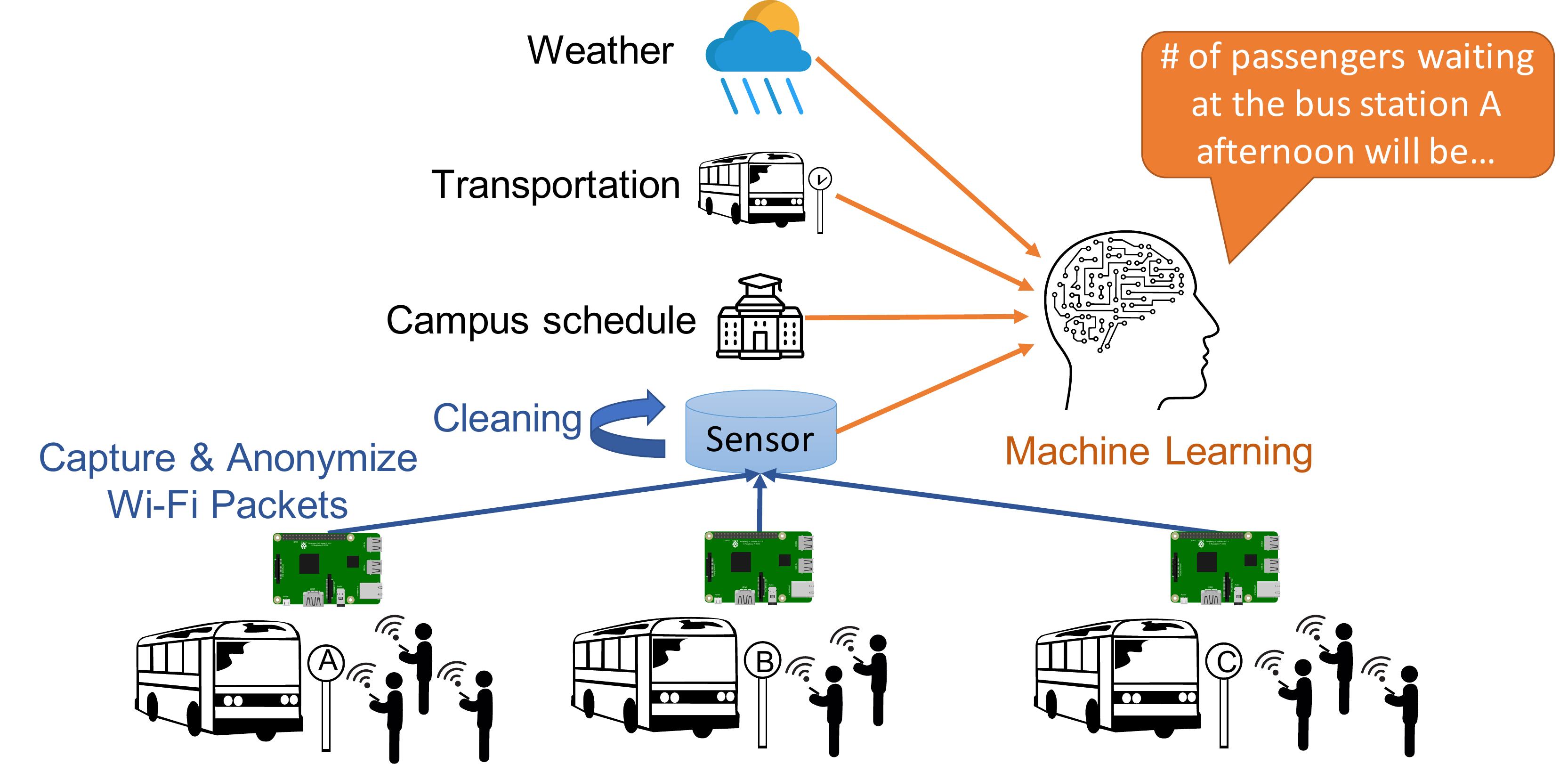}
 \caption{System overview}
 \label{fig:overview}
\end{figure}

\section{Framework}\label{sec:framework}

The proposed system estimates the number of passengers with a machine learning technique taking as an input anonymous and abstracted Wi-Fi capture data at each bus stop and weather data as shown in Figure~\ref{fig:overview}. Some of the Wi-Fi captured data might correlate with the number of persons since they have a Wi-Fi enabled smartphone. On the other hand, we should keep in mind the privacy concerns during capturing Wi-Fi frames, including MAC addresses. Weather data is also informative for estimating the number of passengers. For example, students would not wait for a bus for several minutes on a sunny day. The research question is whether the weather data and the campus schedule data effectively work as an input to machine learning for estimating the number of passengers.

\subsection{System Design and Rational}



\begin{figure}[!b]
\centering
 \includegraphics[width=0.75\linewidth,clip]{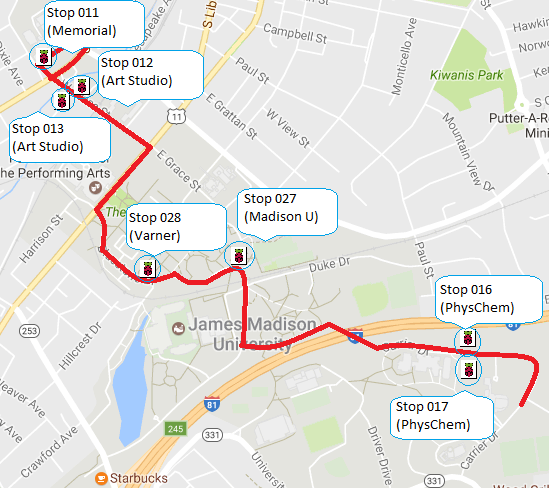}
 \caption{Edge nodes located through the bus route}
 \label{fig:BusRoute}
\end{figure}

The proposed system is based on an IoT cloud computing model. The edge nodes are Raspberry Pi $3$ (located at $7$ different stations as shown in Figure~\ref{fig:BusRoute}) with a monitor mode enabled Wi-Fi interface that captures frames and uploads them to the cloud-based database. Since edge nodes do not require heavy computational need, it works with a solar panel and a rechargeable battery. These edge nodes' main functions are to capture the unencrypted frames, anonymize the MAC addresses with SHA-1, and upload the data to the cloud-based database. The other processes work on the cloud as follows:
\begin{itemize}
\item \textbf{Cleaning}: Filter only the passengers' smartphones frames. Many Wi-Fi frames have noises due to MAC randomization, smartphones passing by a bus stop, and devices in personal vehicles and buildings.

\item \textbf{Learning}: Training the model by learning weather data, Wi-Fi sensor data, and some categorical information for estimating the number of passengers waiting for the bus at the bus stop.

\item \textbf{Estimating}: Estimating the number of passengers from weather information and categorical information.
\end{itemize}
Our system makes sure that only light processing tasks are done on the edge nodes, while the heavy tasks of machine learning and computing are done on the cloud. We installed an IoT node at each bus stop to capture the Wi-Fi frames of passengers' smartphones. There are mainly two ways to set the IoT devices in the field. One is to install them inside the bus (hereafter call it mobile sensing), the other is to place them at each bus stop (fixed point sensing). Estimating the ridership is relatively easy with the mobile sensing against the fixed point sensing because the mobile sensing can track all the passengers inside the bus during driving. However, mobile sensing has no way to know the number of passengers waiting at the bus stop when the bus is not at the bus stop. The number of required devices is the same as the number of buses and bus stops with mobile sensing and fixed point sensing. The fixed point sensing is reasonable in our field because the number of bus stops is less than the buses.

\begin{table}[!ht]
  \begin{center}
    \caption{Weather data}
    \label{table:weatherSpec}
    \begin{tabularx}{\linewidth}{l|X}
      \hline
      Name & Content\\
      \hline
      dt & Time of data calculation, unix, UTC\\
      temp & Temperature\\
      feels\_like & Accounting for the human perception of weather\\
      temp\_min & Minimum temperature among sensors in city\\
      temp\_max & Maximum temperature among sensors in city\\
      pressure & Atmospheric pressure (on the sea level), hPa\\
      sea\_level & Sea Level, meters\\
      grnd\_level & Ground Level, meters\\
      humidity & Humidity, \%\\
      wind\_speed & Wind speed. Unit Default: meter/sec\\
      wind\_deg & Wind direction, degrees (meteorological)\\
      rain\_1h & Rain volume for the last hour, mm\\
      rain\_3h & Rain volume for the last 3h, mm\\
      snow\_1h & Snow volume for the last hour in liquid state, mm\\
      snow\_3h & Snow volume for the last 3h in liquid state, mm\\
      clouds\_all & Cloudiness, \%\\
      weather\_id & Weather condition id\\
      weather\_main & Rain, Snow, Extreme etc.\\
      weather\_description & Weather condition within the group\\
      \hline
    \end{tabularx}
  \end{center}
\end{table}

\subsection{Data Preparation}

\begin{table}[t]
  \begin{center}
    \caption{Raw sensor data}
    \label{table:rawSensorDataSpec}
    \begin{tabular}{l|l}
      \hline
      Name & Type\\
      \hline
      Bus stop & String\\
      Date and time in UTC & YYYY-MM-DD hh:mm:ss\\
      MAC Address & 6 Octet\\
      Signal Strength (dBm) & Integer\\
      \hline
    \end{tabular}
  \end{center}
\end{table}

\begin{table}[t]
  \begin{center}
    \caption{The number of people data}
    \label{table:personsCountSpec}
    \begin{tabular}{l|l}
      \hline
      Name & Type\\
      \hline
      Bus stop & String\\
      Date and time in UTC & YYYY-MM-DD hh:mm:ss\\
      Count & Integer\\
      \hline
    \end{tabular}
  \end{center}
\end{table}

As described in Figure~\ref{fig:overview}, the proposed system employs four types of data for estimating the number of passengers: the weather data (Table~\ref{table:weatherSpec}) obtained from OpenWeather\cite{openweather}, the transportation data (e.g., the locations of the bus stops), the campus schedule (e.g., class schedule), and the number of people data converted from the raw data. Table~\ref{table:rawSensorDataSpec} shows the specification of the raw data. The IoT sensor data is cleaned and transformed into the number of people data (Table~\ref{table:personsCountSpec}) to feed the machine learning.
In the training phase, the machine learning model inputs all the data, while in the testing phase, it does not input the passenger count data since it is the expected output of the model.

\begin{figure}[!hb]
\centering
 \includegraphics[width=\linewidth,clip]{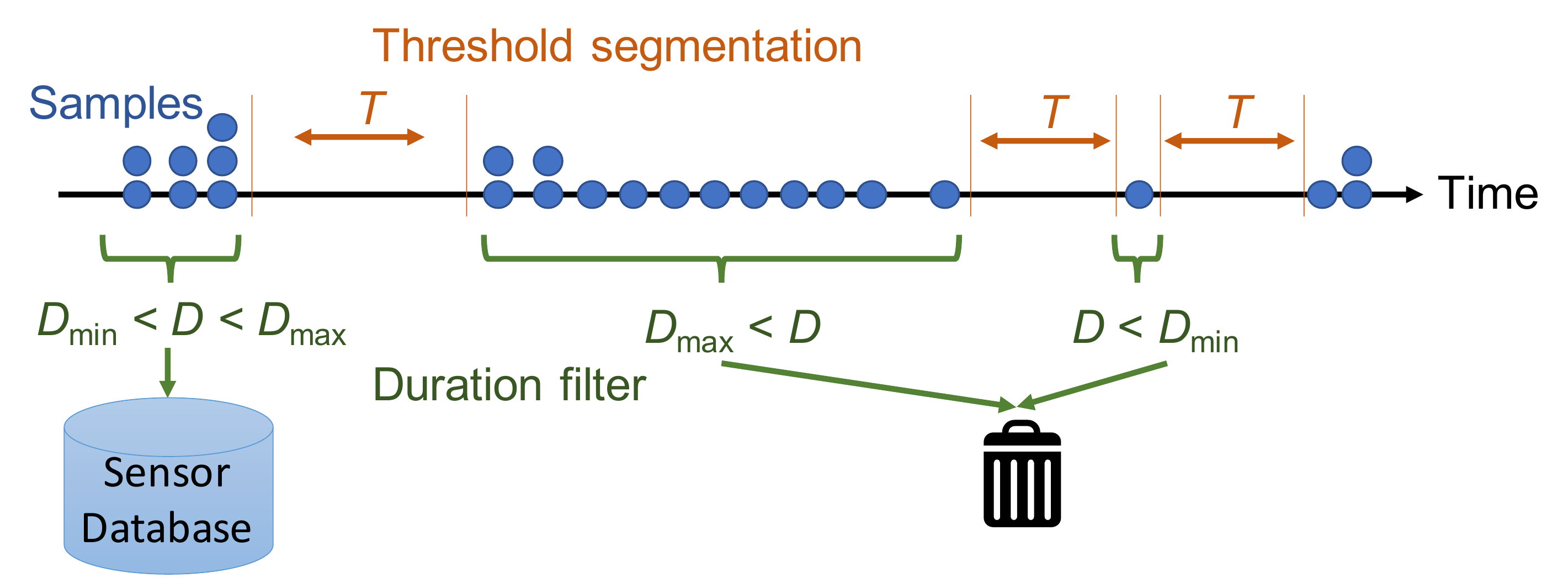}
 \caption{Overview of the segmentation and the filtering}
 \label{fig:durationFilter}
\end{figure}

\begin{figure}[!ht]
 \subfloat[Raw data]{
  \includegraphics[width=0.97\columnwidth,clip]{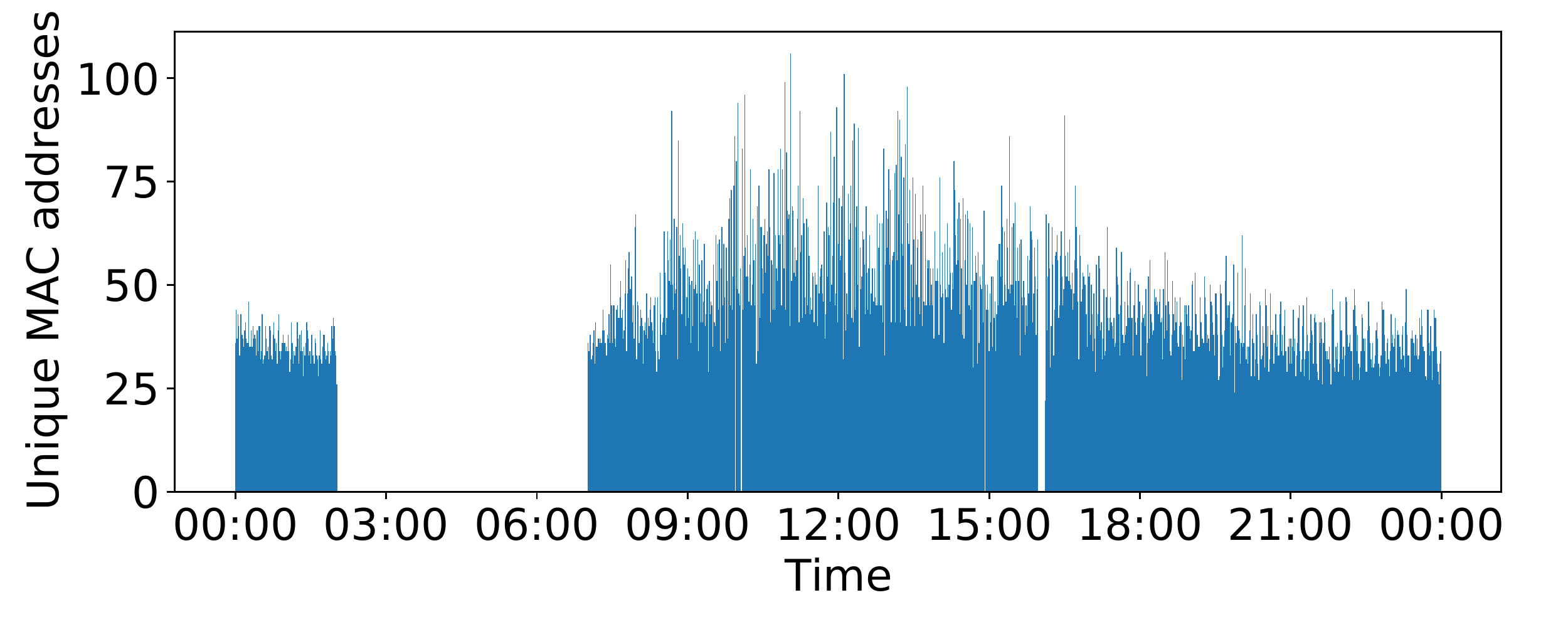}
 }
 
 \subfloat[Cleaned data]{
  \includegraphics[width=0.97\columnwidth,clip]{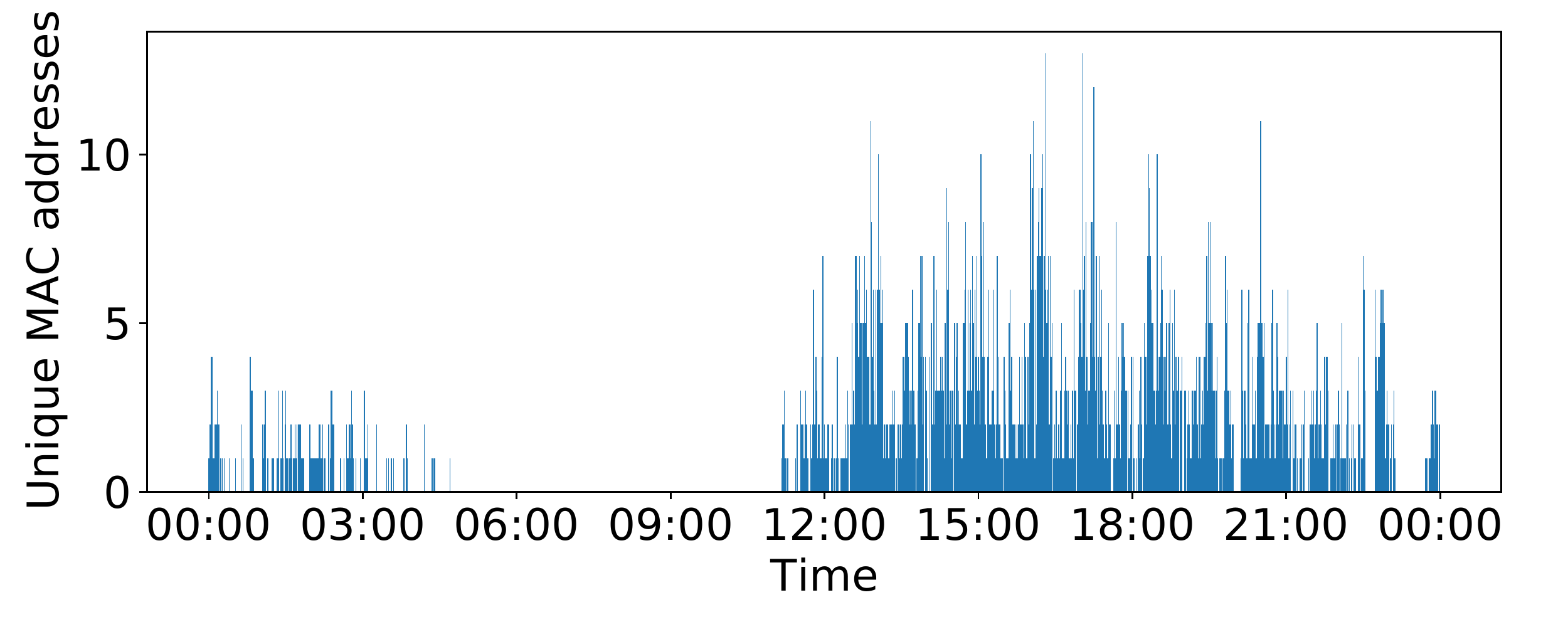}
 }
 \caption{MAC addresses every minutes at 017 (PhysChem) on 4/21}
 \label{fig:countChart}
\end{figure}
We have cleaned the data as follows:
\begin{enumerate}
    \item \textbf{Random MAC Address}: For a privacy measure, iOS 8 and Android 8 started to provide a MAC randomization feature, randomizing a source MAC address when the device is trying to find a Wi-Fi network with a probe request frame. When we identify the devices with the probe request frame, the random MAC address will be noise. We have removed such frames by checking G/L and I/G bits in the MAC address. \footnote{Our data collection was done while iOS versions implementing randomization were not yet released.}

    \item \textbf{MAC Addresses observed at only one bus station}: The passengers' smartphone's Wi-Fi frames should be captured at two or more bus stations. If a MAC address is found at only one bus stop, we assume that a person only passed by the bus stop or a device is installed around the bus stop (e.g., Wi-Fi enabled PC in a building). We know that a passenger phone battery may die while riding the bus; we assume that this case can be ignored with big data.

    \item \textbf{Segmentation and filtering based on the duration}: As shown in Figure~\ref{fig:durationFilter}, samples are split into segments with a given threshold time. If the segment's period is shorter than $D_{\text{min}}$ or longer than $D_{\text{max}}$, it is discarded. The rest of the segments will be stored in the cloud-based database.

    \item \textbf{Unrealistic RSSI}: Usually, the waiting passengers' length is within ten or more meters. In this condition, Received Signal Strength Indication (RSSI) will be in a certain range.
\end{enumerate}
Figure~\ref{fig:countChart} shows the result of cleaning the data of bus stop $017$ (PhysChem) on April \nth{21} in 2017. We used 2 and 30 minutes for $D_{\text{min}}$ and $D_{\text{max}}$, respectively. And we assumed the RSSI range for capturing Wi-Fi frames from smartphones within 10 meters is between -30 dBm and -80 dBm. Against the raw data shows the unrealistic numbers of unique MAC addresses all over the time, the cleaned data shows like a transition of the number of passengers in a day. Though we have not confirmed the cleaned data's correctness due to the lack of the ground truth, it will be no problem because it is generally consistent with the trends we have seen visually at the bus stop.

To prepare machine learning data, we derived features (e.g., academic week) as a week of the semester from the date. Also, we derived the morning/evening feature from time. We further created dummy variables for all values of categorical features (e.g., weather\_description and Bus\_stop) in Tables~\ref{table:weatherSpec} and \ref{table:rawSensorDataSpec}, respectively. For example, the weather\_description having "Rain" and "Cloud" will be True/False values of weather\_description\_Rain, and weather\_description\_Cloud.

\subsection{Machine Learning}
For the Transit Bus Management System, data-driven fleet management strategies empowered by precise models that predict the number of passengers waiting at transit bus stops based on real-time weather conditions are essential. For example, the system could suggest dispatching many buses to a particular route depending on the weather conditions, instead of having near-empty buses some days and full one's others. We highlight the following machine learning techniques:
Linear Regression (LR) analysis is a predictive modeling technique that estimates the relationship between two or more variables. Recall that a correlation analysis does not assume the causal relationship between two variables. 

\begin{equation}
Y_{i}=f(X_{i},\theta)+\epsilon_{i} 
\end{equation}
where for $k$ features and $n$ data points, $i = 1, ..., n$ and $\theta$ = ($\theta_{1}$,...,$\theta{k})^T$ is the vector of the parameters to be estimated. The error $\epsilon_{i}$ are with a mean equal to zero and an unknown $\sigma^2$. 
LR can be used when the features measured have a linear correlation with the dependent variable. Since not all features are having a linear relationship with the passengers' count to be predicted, it is expected that good results cannot be obtained using a plain LR model. 

Classification and regression trees (CART) are binary decision trees. The tree is constructed by splitting the entire data into subsets by using all the independent variables. The goal is to produce terminal leaves that are as homogeneous as possible with respect to the target variable. Regression trees can be notable accurate in the case of nonlinear problems. For every node $t$, $1/n\sum_{i=1}^{n}(Y_i - \bar{Y_i}(t))^2$ is the node sum of squares. In other words, it is the total squared deviations of $Y_i$ in $t$ from their average. The regression tree is formed by splitting the nodes iteratively so that the decrease in R(T) is maximized, where R(T) sums up all the sums of squares within all the nodes.


%

Neural networks (NN) are models in which input features flow through hidden layers towards the output~\cite{hecht1992theory}. The neural network learns new feature spaces by first computing the affine (linear) transformations of the given inputs and then applying a nonlinear function (rectified linear unit ReLU), which will be the next layer's input. This process will continue just before the output layer when a linear transformation will be applied for predicting the hourly passenger demand. NN with no hidden layers can only learn how to solve linearly separable problems. NN with two or more hidden layers capture non-linearity is a natural aspect. NN with three hidden layers is usually considered a deep-neural network. For this study, we expect the performance of a deep neural network (DNN) model that consists of an input layer, three hidden layers, and an output layer to outperform LR, CART, and Neural Networks with one hidden layer (WNN).

\begin{figure}[t]
 \centering
 \includegraphics[width=0.66\linewidth]{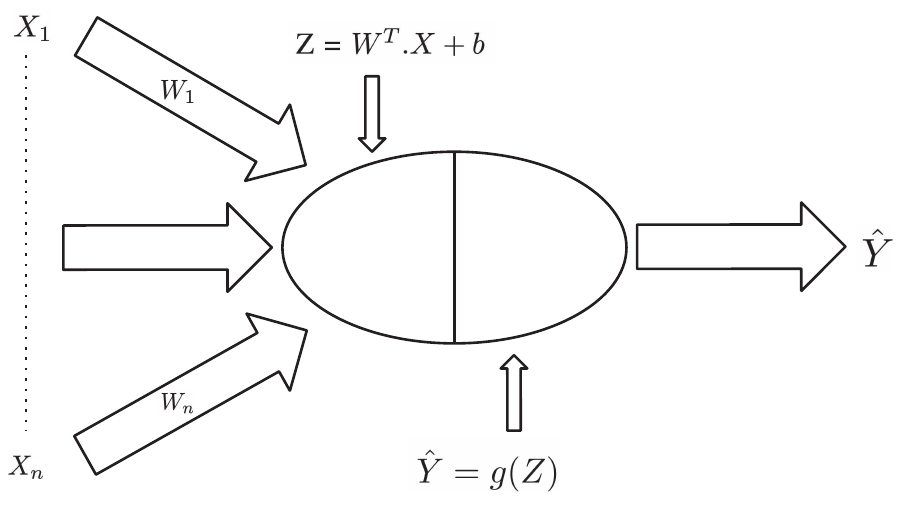}
 \caption{Neural Network Neuron}
 \label{fig:Neuron}
\end{figure}

In the first layer, layer 1 of the NN architecture, each neuron receives a set of X-values (numbered from 1 to $n$) from input vector $X$ and computes the predicted  $\hat{Y}$ value. Vector $X$ contains the weather's value and the bus features, for one example, from the training set. Each node in layer 1 has its own set of parameters, usually referred to as $W$ (column vector of weights) and $b$ (bias), as shown in Figure~\ref{fig:Neuron}. In each iteration, the neuron calculates a weighted average of the vector $X$, based on its current weight vector $W$. Then, it adds bias where weights and bias change during the learning process to minimize prediction error. Finally, the result of this calculation is passed through a ReLU. 

The equation in Figure~\ref{fig:Neuron} used $X$ and $\hat{Y}$, which are the column vector of features and the predicted value, respectively, for a node in layer 1. The $X$ vector is, therefore, layer 0 (input layer). When switching to the general notation for layer $k$, like layer 2 and layer 3, we use $a^{[k-1]}$ and $a^{[k]}$, which are the input to the hidden layer $k$ and the activated output predicted value of layer $k$, respectively (activation function $g$ used is ReLU). Accordingly, for any hidden layer $k$ with $n^{[k]}$ neuron, each neuron performs a similar calculation according to the following equations: 

\begin{equation}
Z_{i}^{[k]}=W_{i}^T .a^{[k-1]} + b_{i}
\end{equation}
\begin{equation}
a_{i}^{[k]}=g^{[k]}(Z_{i}^{[k]})
\end{equation}

The activation function $g$ for any hidden layer is ReLU, except for the output layer, which is a linear function to predict the hourly passenger demand. The model learns through iterations of forward-feeding and backward propagation from the input layer, hidden layers, to output and back. Back-propagation is seen as a common approach where random weights are assigned; the output seen is compared with the test data; the output error is calculated comparing the two (i.e., actual output vs. expected output in MSE loss function).
The layer immediately closer to the output layer adjusts its weights, leading to weight adjustments in the subsequent inner layers until the error rate is reduced~\cite{hecht1992theory}. NN models extract features by weight allocation and weight decay through iterations of forward-feeding and backward propagation techniques. 

\section{Experimental results and discussion} \label{sec:results}

\begin{table}[t]
  \begin{center}
    \caption{Experimental data specification}
    \label{table:expDataSpec}
    \begin{tabular}{l|r}
      \hline
      Name & Value\\
      \hline
      Date & From 2017-04-05 to 2017-05-04\\
      Bus stops & 7\\
      Raw data records & 210,161,203\\
      Cleaned data records & 9,981,883\\
      \hline
    \end{tabular}
  \end{center}
\end{table}

\begin{figure}[t]
\begin{center}
 \includegraphics[width=\linewidth, clip]{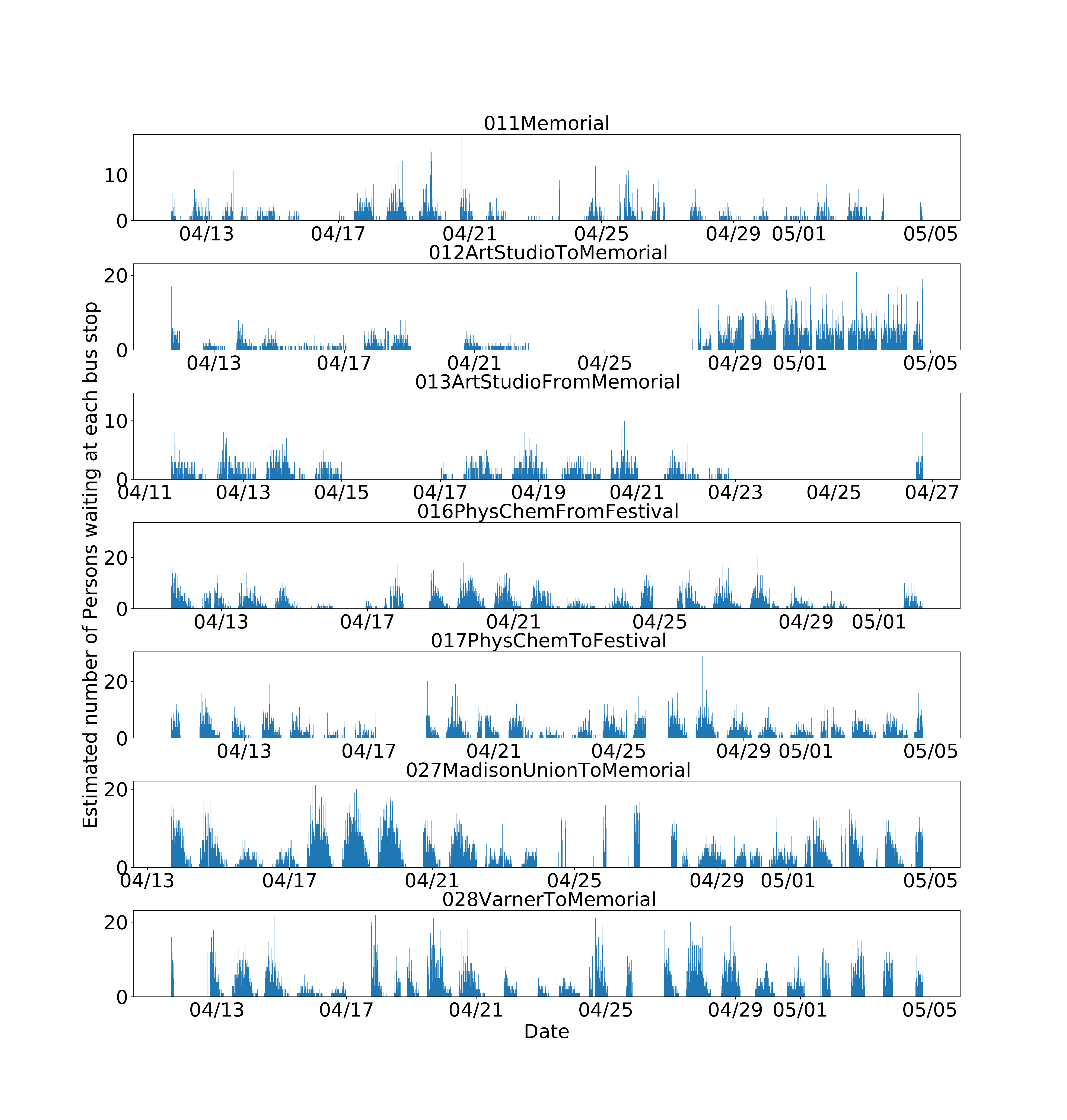}
 \caption{Estimated numbers of passengers waiting at each bus stop with the cleaned data.}
 \label{fig:PersonsEachBusStop}
 \end{center}
\end{figure}

We have collected the experimental data for a month, as shown in Table~\ref{table:expDataSpec}. IoT devices were installed at the seven bus stops shown in Figure~\ref{fig:BusRoute} from April \nth{5} to May \nth{4} in 2017. Figure~\ref{fig:PersonsEachBusStop} shows the estimated numbers of passengers waiting at each bus stop with the cleaned data. The numbers of passengers were around 20.

\begin{figure}[t]
 \centering
 \includegraphics[width=\linewidth]{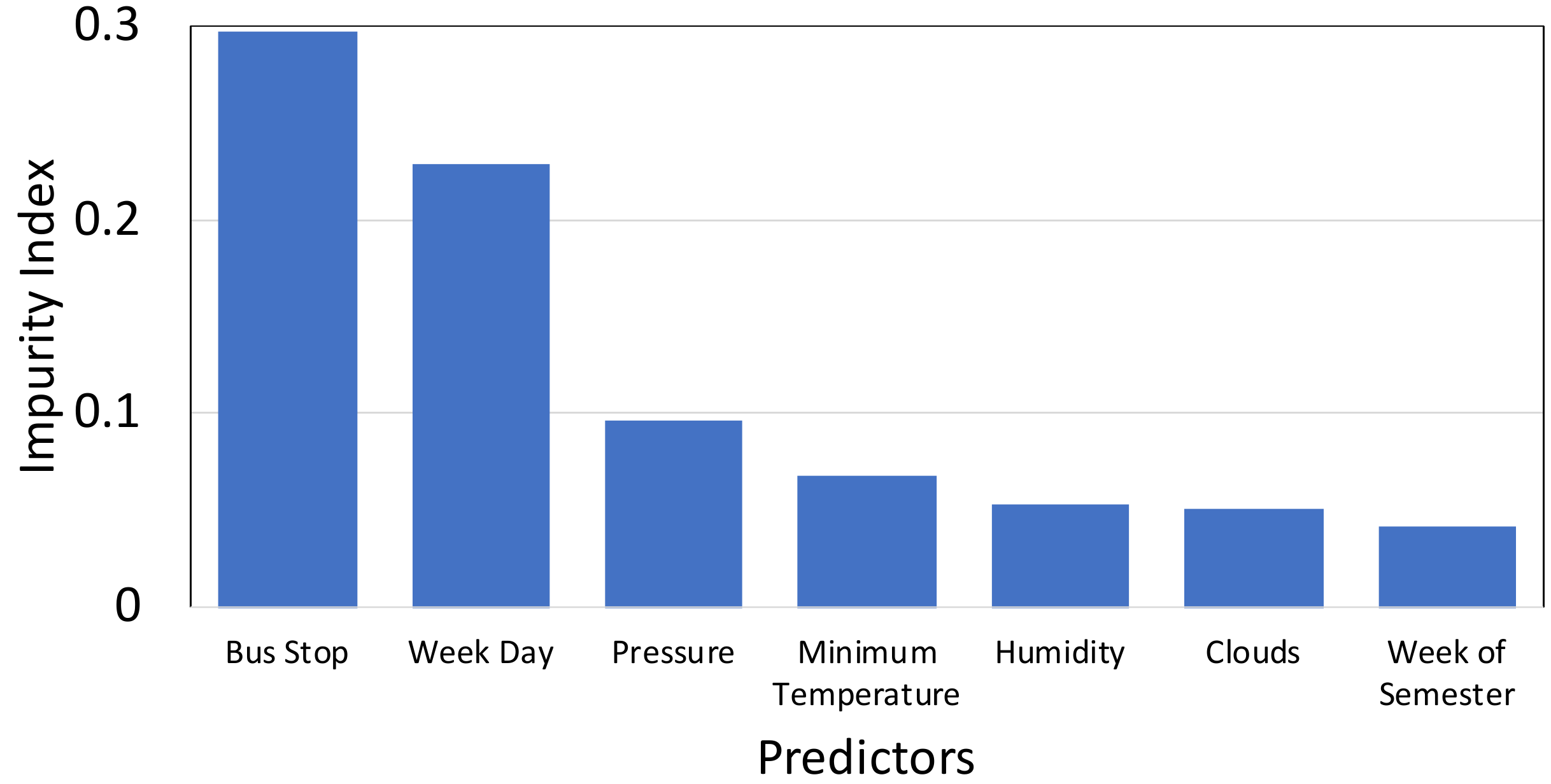}
 \caption{Predictors Impurity Index}
 \label{fig:Impurity}
\end{figure}

The weather predictors (Table~\ref{table:weatherSpec}) and the passenger counts (Table~\ref{table:personsCountSpec}) are recorded hourly and every second, respectively.
We aggregated the passenger counts data on an hourly basis and averaged the reading for the predictors. We then merged the aggregated bus data with the weather data to have our granularity of analysis on the hourly level. We further normalized our data and derived new predictors like weekdays/weekends and morning/evenings. We also extracted the week of the academic semester from the date, which is shown to be informative predictive features, as in Figure~\ref{fig:Impurity}.

After the data preparation and transformation stage, we benchmarked our DNN model (an input layer, three hidden layers, and one output layer) against baseline models: a WNN model with one hidden layer only and traditional LR. Unlike traditional computational intelligence approaches, deep learning predictive performs much better with large datasets. The model learns features to look for and to make better predictions. We split the data randomly to 80\% training and 20\% testing. 

\begin{figure}[t]
 \centering
 \includegraphics[width=\linewidth,clip]{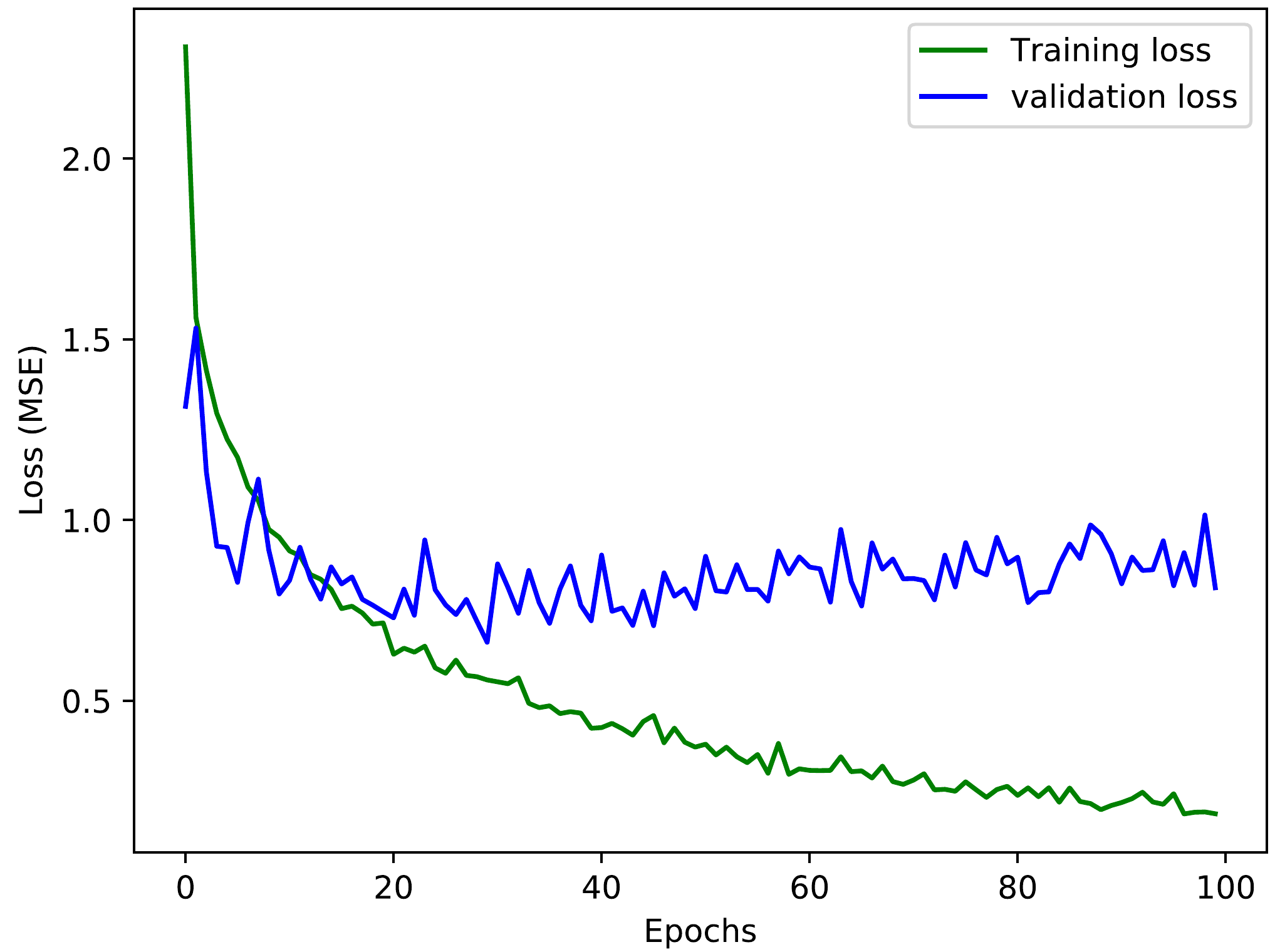}
 \caption{Deep Neural Network MSE}
 \label{fig:deepNNLoss}
\end{figure}

\begin{figure}[t]
 \centering
 \includegraphics[width=\linewidth,clip]{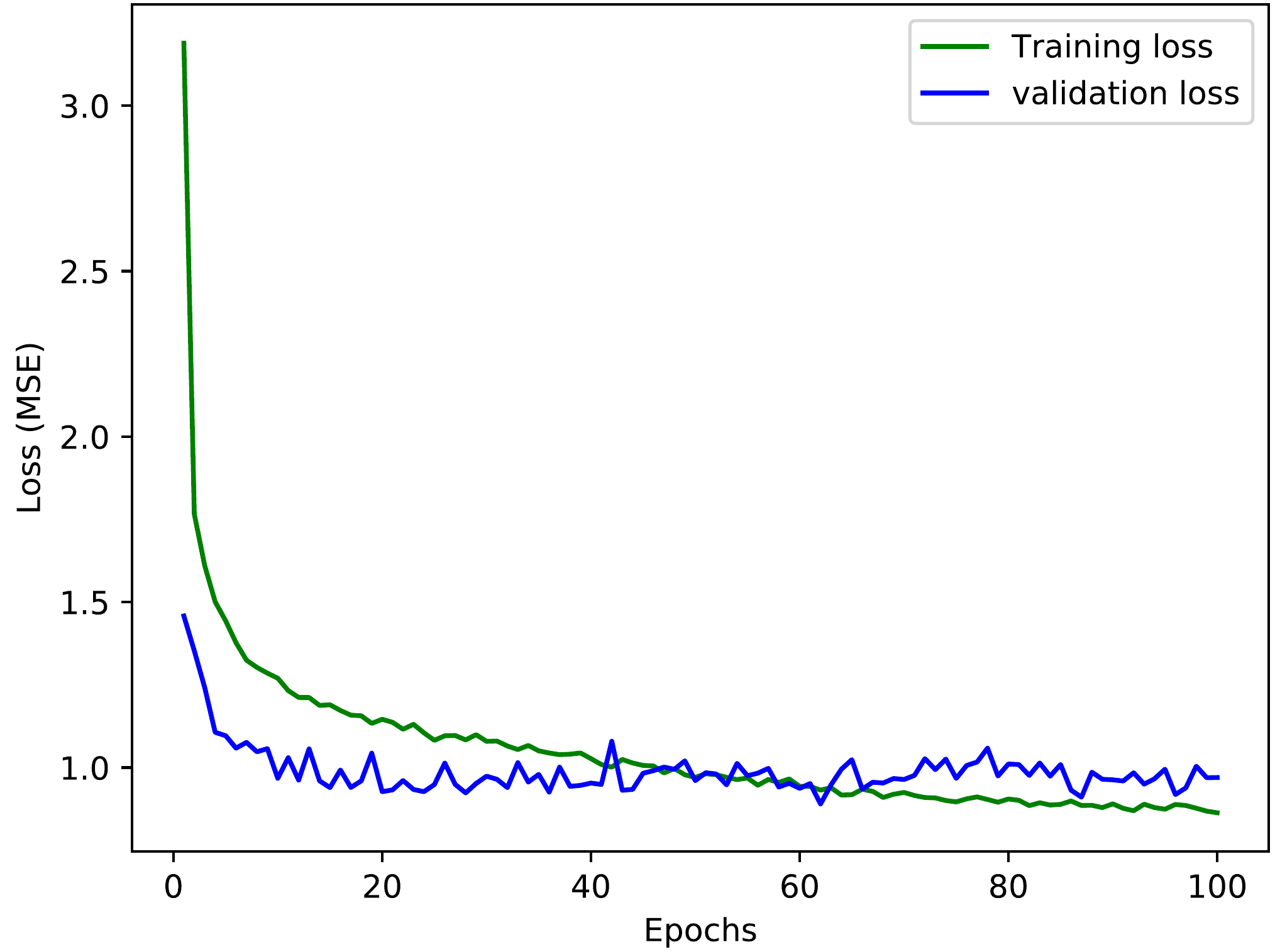}
 \caption{Wide Neural Network MSE}
 \label{fig:wideNNLoss}
\end{figure}

For better prediction, we further fine-tuned the DNN and WNN through another round of random splitting of the training dataset into 80\% training and 20\% validation. We then plotted and analyzed the MSE of the training and validation datasets for WNN and DNN over 100 epochs, as shown in Figures~\ref{fig:deepNNLoss} and ~\ref{fig:wideNNLoss}. 

\begin{figure}[t]
 \centering
 \includegraphics[width=\linewidth,clip]{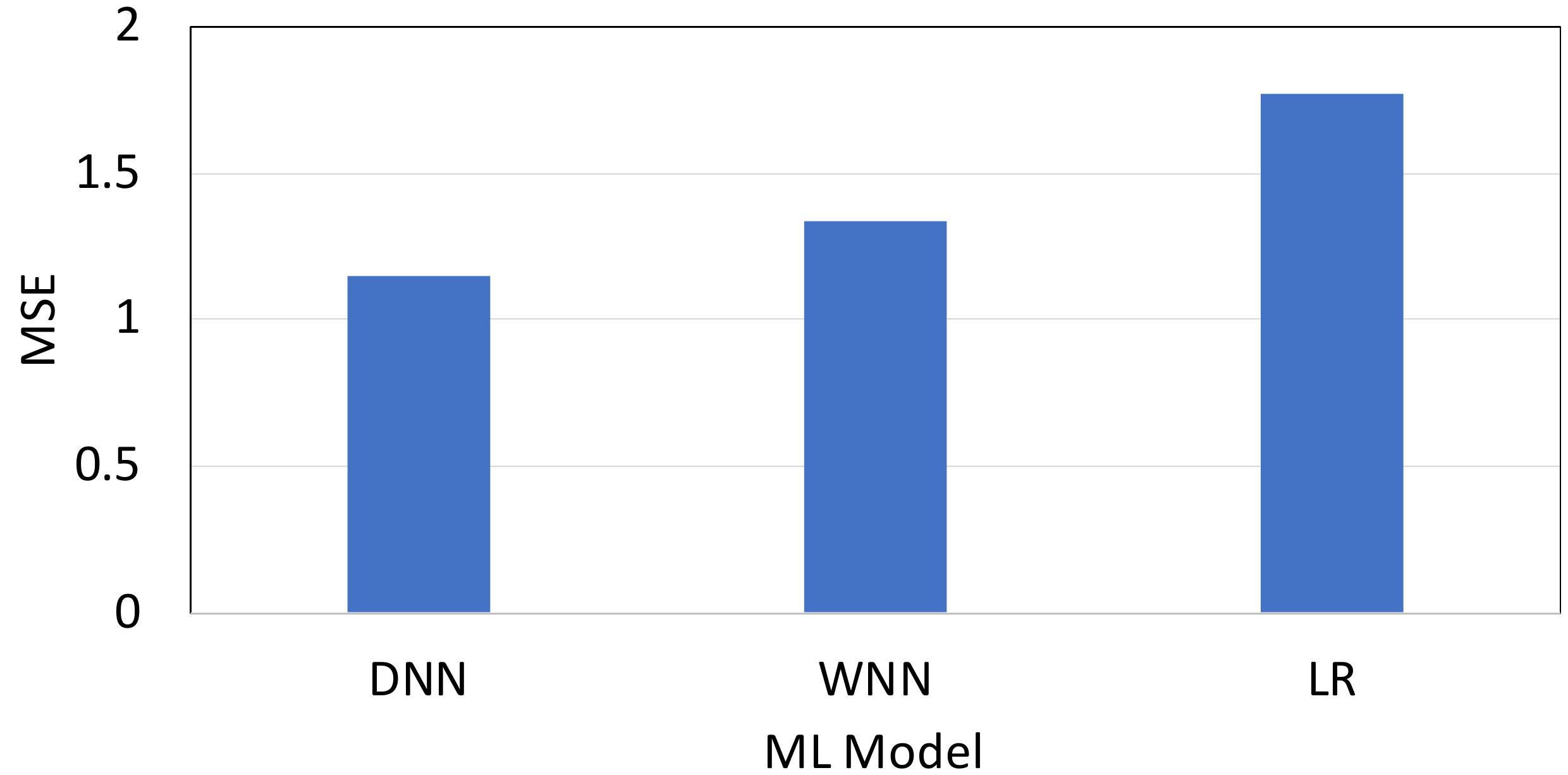}
 \caption{Mean Squared Error (MSE) of Medels.}
 \label{fig:MSE}
\end{figure}

As a final step, and after fine-tuning the parameters of NN models using the validation dataset, we compared the DNN model against the two baseline models using the testing dataset. The MSE for the testing dataset for the DNN model was $1.15$, compared to $1.34$ and $1.77$ for WNN and LR models, respectively, as shown in Figure~\ref{fig:MSE}. We offer empirical evidence on DNN models' effectiveness for predicting the hourly number of passengers waiting in bus stations by achieving  $35\%$ and $14\%$ better prediction for the DNN model over LR and WNN models, respectively. 



Furthermore, to study the influence of our dataset’s informative features, we use an ensemble of gradient boosting decision trees. A benefit of using ensembles of decision tree methods like gradient boosting is that they can automatically estimate feature importance from a trained predictive model. Decision trees provide more information about the relationships of the features than a standard correlation index. We rank the predictive power of the input features, and we find that there are seven most decisive features for partitioning the decision tree to predict the number of passengers waiting at the bus stop. The seven features as shown in Figure~\ref{fig:Impurity} are Bus Stop, Week Day, Pressure, Minimum Temperature, Humidity, Clouds, and Week of Semester.
It summarizes the reduction in the impurity index over all trees when a particular feature is pointed during the trees’ internal space partition over several epochs.


\section{Conclusion}\label{sec:final}



This paper studied the correlation between the passengers waiting at bus stations and the weather conditions using a deep learning algorithm. This paper gives empirical evidence on the importance of incorporating predictive modeling in intelligent transportation to maximize fleet utilization. We highlighted the importance of applying deep learning models to precisely predict the number of passengers waiting at bus stops in intelligent transportation systems. This research also leveraged a broad set of features from IoT devices like smartphone Wi-Fi data in conjunction with detailed weather information. Our results show four of the top seven decisive features (Pressure, Minimum Temperature, Humidity, and Clouds) were related to temperature and weather.  It gives empirical evidence on the importance of weather information in influencing riders’ behavior and, hence, better predicting the number of passengers waiting at the bus stops. 

We studied only one month of campus-level data in 2017. In future work, data collection could take place yearly and on the city scale to confirm the generalization of the predictive model's performance. A limitation of this study is that we did not take into consideration any unusual event like a pandemic. In future work, we will incorporate data of 2020 and investigate more informative features. Among the features to be explored are the government's restriction levels (stay-at-home, business is open with restrictions or no restrictions, etc.) The study should also incorporate features like the number of daily reported cases reported in the city and lockdown dates to improve the prediction of the number of passengers waiting at bus stops during the COVID-19 pandemic.


\section*{ACKNOWLEDGMENT}
This work is funded through a 4-VA research grant (https://4-va.org) and JSPS KAKENHI Grant Numbers JP20K11789, JP20H04183.

\bibliographystyle{./bibliography/IEEEtran}
\bibliography{./bibliography/IEEEabrv,./bibliography/IEEEexample,./bibliography/reference.bib}

\end{document}